\documentclass{article} % For LaTeX2e
\usepackage[preprint]{colm2026_conference}

\usepackage{microtype}
\usepackage{hyperref}
\usepackage{url}
\usepackage{booktabs}
\usepackage{multirow}
\usepackage{amsmath}
\usepackage{mathtools}
\usepackage{subcaption}
\usepackage{enumitem}
\usepackage{wrapfig}
\definecolor{darkblue}{rgb}{0, 0, 0.5}
\hypersetup{colorlinks=true, citecolor=darkblue, linkcolor=darkblue, urlcolor=darkblue}

\title{Distributed Multi-Layer Editing for Rule-Level Knowledge in Large Language Models}

\author{
Yating Wang$^{1}$, Wenting Zhao$^{2}$, Yaqi Zhao$^{1}$, Yongshun Gong$^{1}$, Yilong Yin$^{1}, $Haoliang Sun$^{1}$ \\
$^{1}$School of Software, Shandong University, Jinan, China \\
$^{2}$Salesforce AI Research
}

\begin{document}

\ifcolmsubmission
\linenumbers
\fi

\maketitle

\begin{abstract}
% Model editing aims to efficiently update large language models with new information. While the prevailing ``locate-then-edit'' paradigm has demonstrated success in editing fact-level knowledge, it remains limited for rule-level knowledge. A mathematical rule can be expressed through multiple interdependent forms: formulas, descriptions, and instances, which require consistent updates across representations. In this work, we first construct \textbf{RuleEdit-200}, a dataset of 200 rules with three aligned representations, and then conduct a fine-grained causal tracing analysis to examine how different rule forms are organized across transformer layers in large language models. Our analysis reveals a consistent form-specific pattern: formulas and descriptions are mainly associated with earlier layers, while instances rely more on middle layers. This finding suggests that rule knowledge is not uniformly localized, and that effective rule editing should follow this internal organization rather than rely on a single localized intervention. Guided by this observation, we propose \textbf{Distributed Multi-Layer Editing} (DMLE), a framework that applies a shared early-layer edit to formulas and descriptions and a separate middle-layer edit to instances. Experiments across multiple models show that DMLE achieves more robust and consistent rule-level knowledge editing than existing methods.

Large language models store not only isolated facts but also rules that support reasoning across symbolic expressions, natural language explanations, and concrete instances. Yet most model editing methods are built for fact-level knowledge, assuming that a target edit can be achieved through a localized intervention. This assumption does not hold for rule-level knowledge, where a single rule must remain consistent across multiple interdependent forms.
We investigate this problem through a mechanistic study of rule-level knowledge editing. To support this study, we extend the RuleEdit benchmark from 80 to 200 manually verified rules spanning mathematics and physics. Fine-grained causal tracing reveals a form-specific organization of rule knowledge in transformer layers: formulas and descriptions are concentrated in earlier layers, while instances  are more associated with middle layers. These results suggest that rule knowledge is not uniformly localized, and therefore cannot be reliably edited by a single-layer or contiguous-block intervention.
Based on this insight, we propose Distributed Multi-Layer Editing (DMLE), which applies a shared early-layer update to formulas and descriptions and a separate middle-layer update to instances. While remaining competitive on standard editing metrics, DMLE achieves substantially stronger rule-level editing performance. On average, it improves instance portability and rule understanding by 13.91 and 50.19 percentage points, respectively, over the strongest baseline across GPT-J-6B, Qwen2.5-7B, Qwen2-7B, and LLaMA-3-8B. The code is available at https://github.com/Pepper66/DMLE.

\end{abstract}

\section{Introduction}

Model editing aims to update large language models (LLMs) with fresh, corrected, or domain-specific knowledge without the high cost of full retraining \citep{mitchell2021fast, meng2022locating, meng2022mass, fang2024alphaedit, li2025reinforced, deng2025everything}. Most existing methods, however, focus on \emph{fact-level} knowledge, where the editing target is typically a single factual association. In this setting, influential locate-then-edit methods such as ROME and MEMIT assume that a target edit can be achieved through a localized intervention on the model's parameters \citep{meng2022locating, meng2022mass}. While effective for many factual edits, it remains unclear whether this assumption holds for more structured knowledge.

Beyond isolated facts, LLMs also encode rules that support mathematical and scientific reasoning. Unlike a single fact tuple, a rule must remain consistent across multiple interdependent forms, including formulas, descriptions, and instances. Rule-level editing is therefore more challenging than standard fact editing: existing approaches often fail to propagate edits coherently across these forms \citep{zhangruleedit}, suggesting that rule knowledge might not be as localized as facts. Instead, different forms may rely on distinct internal computations, making the core assumption behind localized methods mismatched to rule-level knowledge.

%In this work, we address this question from a mechanistic perspective. We ask how different forms of the same rule are organized across transformer layers, and whether this internal organization can explain the limitations of existing editors. To support this study, we construct \textsc{RuleEdit-200}, a dataset of 200 manually verified rules spanning mathematics and physics, each paired with aligned formula, description, and instance forms. We then perform fine-grained causal tracing to measure how strongly different layers contribute to each form of rule knowledge.
In this work, we address this question from a mechanistic perspective. We ask how different forms of the same rule are organized across transformer layers, and whether this internal organization can explain the limitations of existing editors. To support this study, we extend the \textsc{RuleEdit} benchmark \citep{zhangruleedit} from 80 to 200 manually verified rules spanning mathematics and physics, with each rule paired with aligned formula, description, and instance forms. We then perform fine-grained causal tracing to measure how strongly different layers contribute to each form of rule knowledge.

\begin{figure}[t]
    \centering
    \includegraphics[width=0.85\linewidth]{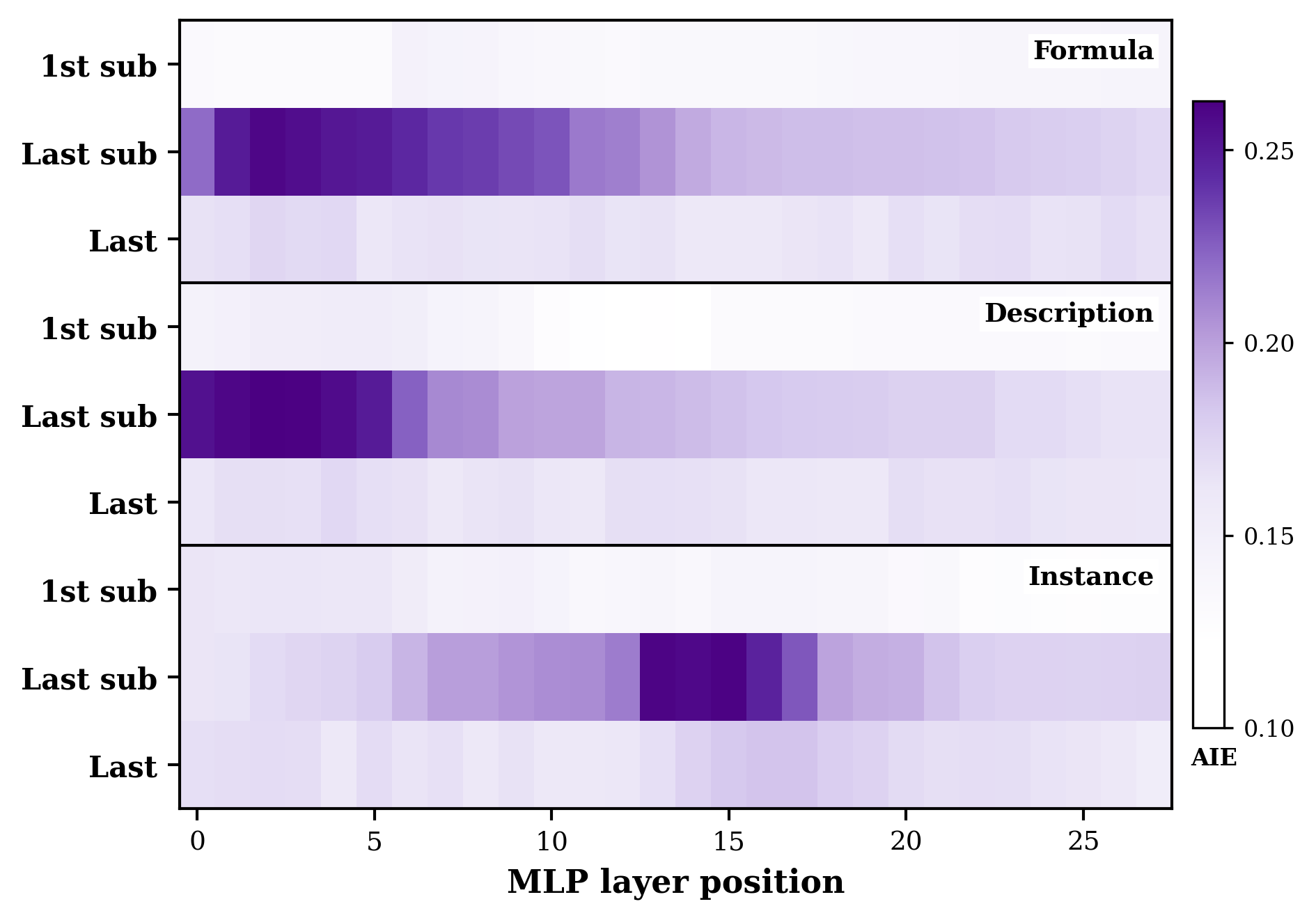}
    \caption{Causal tracing heatmaps on GPT-J-6B. The heatmap shows the average indirect effect (AIE) across MLP layers and token positions for formula, description, and instance. \textit{1st sub} and \textit{Last sub} denote the first and last subject tokens, and \textit{Last} the final prompt token. \textit{Formula} and \textit{description} peak in earlier layers, whereas \textit{instance} peaks in middle layers.}
    \label{fig:causal_tracing_gptj}
\end{figure}

Our analysis reveals a clear \emph{form-specific layer-wise organization} rather than a single localized storage site. This form-specific separation is already visible in Figure~\ref{fig:causal_tracing_gptj}, which shows that on GPT-J-6B, formulas and descriptions exhibit their strongest indirect effects in earlier layers, whereas instances peak in middle layers. This contrast suggests that different forms of the same rule are supported by different parts of the network, rather than by a single shared editing location. As a result, editing only a single layer, or even a small contiguous block of layers, is unlikely to update the full rule coherently across forms.

Motivated by this observation, we propose \textbf{Distributed Multi-Layer Editing (DMLE)}, a rule-level editing framework built on MEMIT. Instead of applying a uniform intervention to all forms, DMLE performs a shared edit for formulas and descriptions in early layers, and a separate edit for instances in middle layers. In this way, DMLE aligns its intervention with the observed internal organization of rule knowledge, enabling form-specific updates at the layers most relevant to each form.
Experiments on GPT-J-6B, Qwen2-7B, Qwen2.5-7B, and LLaMA-3-8B show that DMLE remains competitive on standard editing metrics while substantially improving rule-level editing performance over strong baselines. More broadly, these results suggest that understanding how rule knowledge is internally organized can directly inform more effective editing methods for structured knowledge in LLMs.

%Our contributions are threefold. First, we construct \textsc{RuleEdit-200}, a dataset of 200 rules with aligned formulas, descriptions, and instances, yielding 600 samples for controlled rule-level analysis and editing. Second, we study rule-level knowledge editing from a mechanistic perspective and show through fine-grained causal tracing that different forms of the same rule exhibit a clear layer-wise organization in LLMs. Third, guided by these findings, we propose \textbf{DMLE}, a form-specific editing framework that improves coherent rule updates across multiple base models.
In summary, our work makes three contributions: we extend \textsc{RuleEdit} from 80 to 200 manually verified rules with aligned formulas, descriptions, and instances to support controlled rule-level analysis and editing; we show through fine-grained causal tracing that different rule forms exhibit a clear layer-wise organization in LLMs; and we propose \textbf{DMLE}, a form-specific editing framework that enables more coherent rule updates across base models.

\section{Related Work}

\textbf{Model Editing in LLMs.}
Model editing aims to update specific knowledge in LLMs without full retraining \citep{mitchell2021fast, meng2022locating, meng2022mass, hartvigsen2023aging, fang2024alphaedit, zhang2024locate, huang2024can, yu2024melo, wang2024knowledge, gu2024model, li2025reinforced, deng2025everything, lu2025knowledge, jiang2025anyedit, wei2025mlake}. Most existing methods follow a locate-then-edit paradigm \citep{meng2022locating}, first identifying the model components responsible for the target knowledge and then applying localized parameter updates. Representative approaches include ROME \citep{meng2022locating}, which performs rank-one updates to localized factual associations, and MEMIT \citep{meng2022mass}, which extends this idea to the simultaneous editing of multiple knowledge associations through closed-form weight updates. Despite their success on fact-level editing, these methods generally assume that knowledge can be modified through a single layer or a small contiguous block of layers. In this work, we revisit this assumption in the context of rule-level knowledge editing. AlphaEdit \citep{fang2024alphaedit} offers an alternative approach, using null-space constraints to enable more precise knowledge editing across different parts of the model.

\textbf{Rule-level Knowledge Editing.}
Most prior work on knowledge editing focuses on fact-level knowledge, whereas rule-level knowledge introduces additional challenges due to its structured and multi-form nature. \citet{zhangruleedit} formalize this setting by decomposing each rule into three aligned forms: formula, description, and instance. This benchmark highlights the importance of cross-form consistency and shows that existing editing methods struggle to generalize edits coherently across forms. While this line of work clearly establishes rule-level editing as a challenging problem, the underlying reason for this difficulty remains largely unexplored. Our work addresses this gap by analyzing how different forms of rule knowledge are internally organized in LLMs.

\textbf{Knowledge Localization and Causal Tracing.}
Research on Transformer interpretability has provided important insights into how knowledge is stored and processed in LLMs. \citet{geva2021transformer} show that feed-forward networks (FFNs) in Transformer layers can function as key-value memory modules, providing early evidence for internal knowledge storage. Logit Lens \citep{nostalgebraist2020logit} enables layer-wise probing by projecting intermediate representations into the vocabulary space. \citet{meng2022locating} further introduce causal tracing to localize fact-level knowledge, establishing a practical framework for identifying where model predictions causally depend on internal states. We adopt causal tracing in a different setting: rather than localizing isolated facts, we use it to study the layer-wise organization of multi-form rule knowledge, and to guide the design of rule-level editing.

\section{RuleEdit-200: A Rule-Level Editing Dataset}

To evaluate the effectiveness and consistency of rule-level knowledge editing, we construct RuleEdit-200, a benchmark that extends the framework of \citet{zhangruleedit}. While their dataset contains 80 geometry-related rules, RuleEdit-200 broadens the scope to 200 distinct rules and 600 corresponding samples. Unlike standard fact-level editing benchmarks based on simple fact tuples, RuleEdit-200 is designed to capture the multi-form nature of rules in mathematics and physics. The full dataset are provided in the supplementary material.

\noindent\textbf{Knowledge Sourcing and Domain Coverage.}
We collect 200 fundamental rules from two primary sources: \textit{Wikipedia Mathematics} \citep{wikipedia_math} and introductory physics materials \citep{openstax_physics}. These rules span a range of mathematical and physical concepts, with representative examples drawn from areas such as algebra, geometry, classical mechanics, and electricity. 

\noindent\textbf{Counterfactual Generation and Multi-Form Structure.}
For each rule, we construct a counterfactual target to simulate a rule update scenario. Specifically, we use Gemini \citep{gemini2023} to generate modified rule statements that alter the original rule while keeping the statement natural and well-formed (e.g., changing the geometric mean from $\sqrt(a\times b)$ to $\sqrt(a+b)$). This design ensures that correct predictions depend on the edited rule rather than the model's prior knowledge.

Following the evaluation setting of \textit{Rule-Edit} \citep{zhangruleedit}, each rule is structured into three complementary forms that capture different aspects of rule knowledge. Each editing case therefore contains three components:

% \begin{itemize}
% \item \textbf{Symbolic Formula:} the formal mathematical expression representing the rule (e.g., $\sqrt(a+b)$).
% \item \textbf{Natural Language Description:} a natural language statement that conveys the meaning of the rule (e.g., ``the square root of the sum of the numbers.'').
% \item \textbf{Numerical Instance:} a concrete numerical application of the rule (e.g., $\sqrt(16+9)=5$).
% \end{itemize}
\begin{itemize}[leftmargin=1.5em, itemsep=0.2em, topsep=0.2em, parsep=0pt, partopsep=0pt]
\item \textbf{Symbolic Formula:} the formal mathematical expression representing the rule (e.g., $\sqrt{a + b}$).
\item \textbf{Natural Language Description:} a natural language statement that conveys the meaning of the rule (e.g., ``the square root of the sum of the numbers.'').
\item \textbf{Numerical Instance:} a concrete numerical application of the rule (e.g., $\sqrt{16+9}=5$).
\end{itemize}

This multi-form structure allows us to evaluate whether an edit updates the rule consistently across formulas, descriptions, and instances.

%All generated samples are manually reviewed by a qualified annotator with a bachelor's degree and CET-6-level English proficiency. The review process includes correcting grammatical errors, improving unclear or unnatural wording, verifying cross-form consistency among the formula, description, and numerical instance, and ensuring that each numerical instance is derived from the edited rule rather than the original one. Samples with ambiguous content or semantic misalignment are further revised before inclusion in the final dataset.Additional details on dataset construction, including the sourcing process, counterfactual generation, and quality control, are provided in Appendix~\ref{appendix:dataset_construction}.

All generated samples are manually reviewed by an annotator with a bachelor's degree, strong English proficiency, and familiarity with the task setting. The review process includes correcting grammatical errors, improving unclear or unnatural wording, verifying cross-form consistency among the formula, description, and numerical instance, and ensuring that each numerical instance is derived from the edited rule rather than the original one. Samples with ambiguous content or semantic misalignment are further revised before inclusion in the final dataset. Additional details on dataset construction, including the sourcing process, counterfactual generation, and quality control, are provided in Appendix~\ref{appendix:dataset_construction}.

\section{Causal Tracing for Rule Knowledge}

Rule-level editing aims to update generalizable rule knowledge in LLMs while maintaining consistency across multiple related forms. We represent each rule as a tuple $R=(s, f, d, i)$, where $s$ denotes the subject, and $f, d,$ and $i$ represent the formula, description, and instance forms, respectively. The objective is to update the model such that the edited rule is reflected consistently across all three forms. Unlike standard fact editing, which focuses on a single factual association, rule-level editing requires coherent updates across the symbolic, descriptive, and instantiated representations of the underlying rule.

\subsection{Layer-wise Causal Tracing}

To understand how rule knowledge is organized in LLMs, we perform layer-wise causal tracing \citep{meng2022locating}, focusing on MLP states. This choice is motivated by prior work showing that feed-forward layers act as key-value memories and play a central role in knowledge storage \citep{geva2021transformer, dai2022knowledge}.

\paragraph{Tracing Data Construction.}
To support causal tracing, we construct a dedicated tracing set based on \textsc{RuleEdit-200}. For each rule, we extract the subject and its corresponding prompt--target pairs, reformulating them into a unified tracing format. To ensure sufficient coverage for the analysis, we replicate and shuffle the resulting pairs, yielding 1,000 samples for each form. Prompts are rewritten using templates that explicitly include the rule subject, while unnecessary filler text is removed. Form-specific targets are defined as: the final symbolic expression for the \textit{formula} form, a concise verb-led phrase for the \textit{description} form, and the final answer for the \textit{instance} form. This refinement reduces linguistic variability and provides clearer localized signals for identifying causally relevant layers. Detailed construction procedures are provided in Appendix~\ref{appendix:tracing_data}.

\paragraph{Tracing Method.}

We build on the causal intervention framework of \citet{meng2022locating} and adapt it to the rule-level setting. Instead of localizing the storage site of a single fact, we trace how formula, description, and instance forms of the same rule contribute across layers. For each prompt--target pair $(x, y)$ associated with a rule form, we first run the model on the original input and obtain the clean target probability $p(y \mid x)$. We then construct a corrupted input $\tilde{x}$ by perturbing the input embeddings of the subject tokens. Starting from this corrupted run, we restore the hidden state at layer $l$ to its clean counterpart and measure how much the target probability is recovered:
\[
r_l = p\bigl(y \mid \tilde{x};\, h_l \leftarrow h_l^{\mathrm{clean}}\bigr) - p\bigl(y \mid \tilde{x}),
\]
where $h_l^{\mathrm{clean}}$ denotes the hidden state at layer $l$ from the clean run.

We perform tracing separately for formula, description, and instance prompts, each paired with its corresponding target. This produces three form-specific layer-wise profiles that reveal how different layers causally support different forms of the same rule. Layers with larger $r_l$ are regarded as making a stronger causal contribution to the corresponding form. These profiles provide the basis for our analysis of rule knowledge organization and the identification of the layer groups for subsequent editing.

\paragraph{Tracing Setup.}

For each prompt, we perform three runs: a clean run, a corrupted run, and a corrupted-with-restoration run \citep{meng2022locating}. We conduct this analysis on four widely used autoregressive language models: GPT-J-6B \citep{wang2021gpt}, Qwen2-7B \citep{yang2024qwen2technicalreport}, Qwen2.5-7B \citep{hui2024qwen25codertechnicalreport}, and LLaMA-3-8B \citep{grattafiori2024llama3herdmodels}.

In the corrupted run, we add Gaussian noise $\epsilon \sim \mathcal{N}(0,\nu)$ to the input embeddings of subject tokens. We set \(\nu=0.025\) for GPT-J-6B, Qwen2-7B, and Qwen2.5-7B, and \(\nu=0.027\) for LLaMA-3-8B to induce a comparable level of corruption across models.

\begin{figure*}[t]
    \centering
    \begin{subfigure}[t]{0.32\textwidth}
        \centering
        \includegraphics[width=\textwidth]{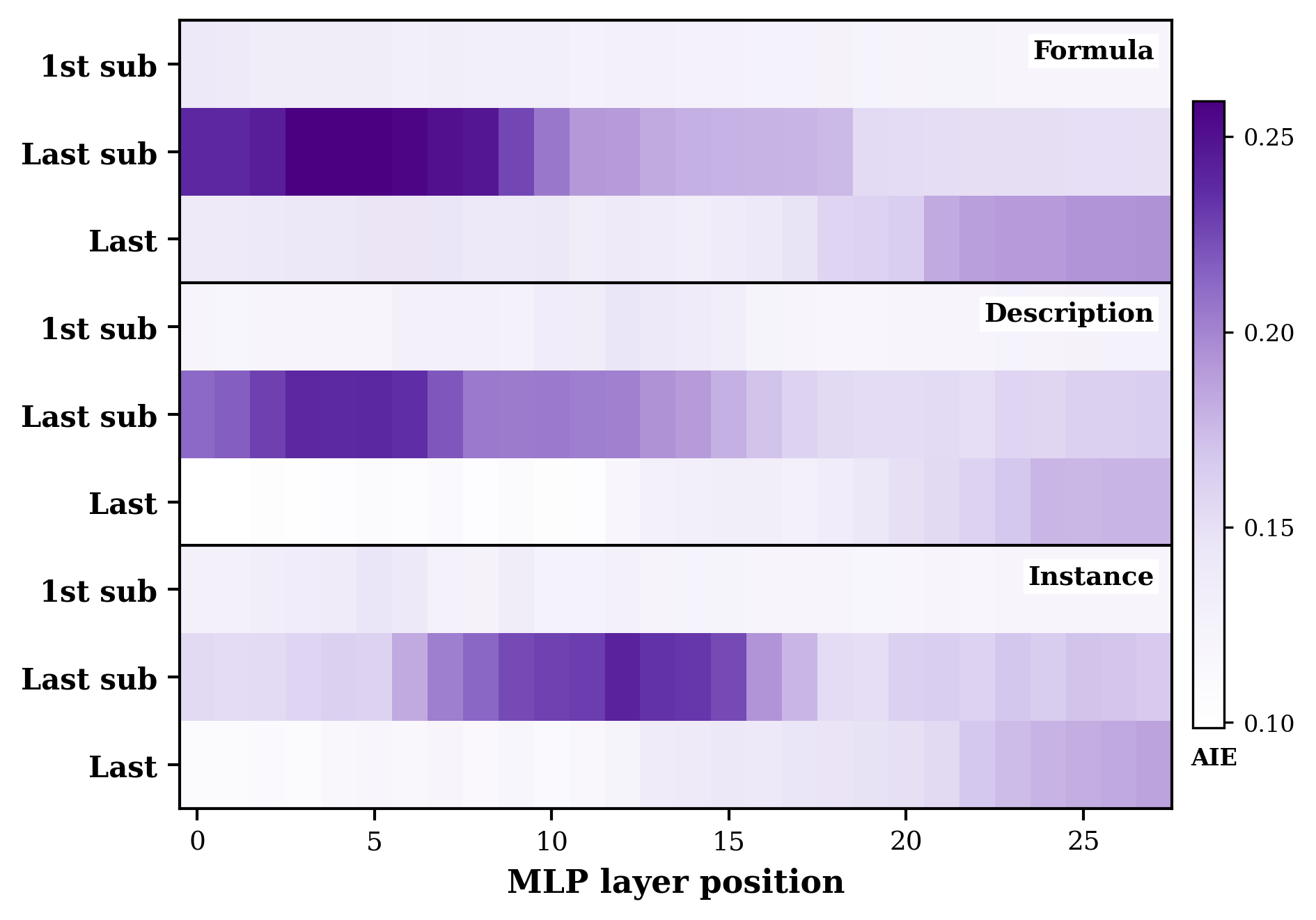}
        \caption{Qwen2-7B}
        \label{fig:causal_tracing_qwen2}
    \end{subfigure}
    \hfill
    \begin{subfigure}[t]{0.32\textwidth}
        \centering
        \includegraphics[width=\textwidth]{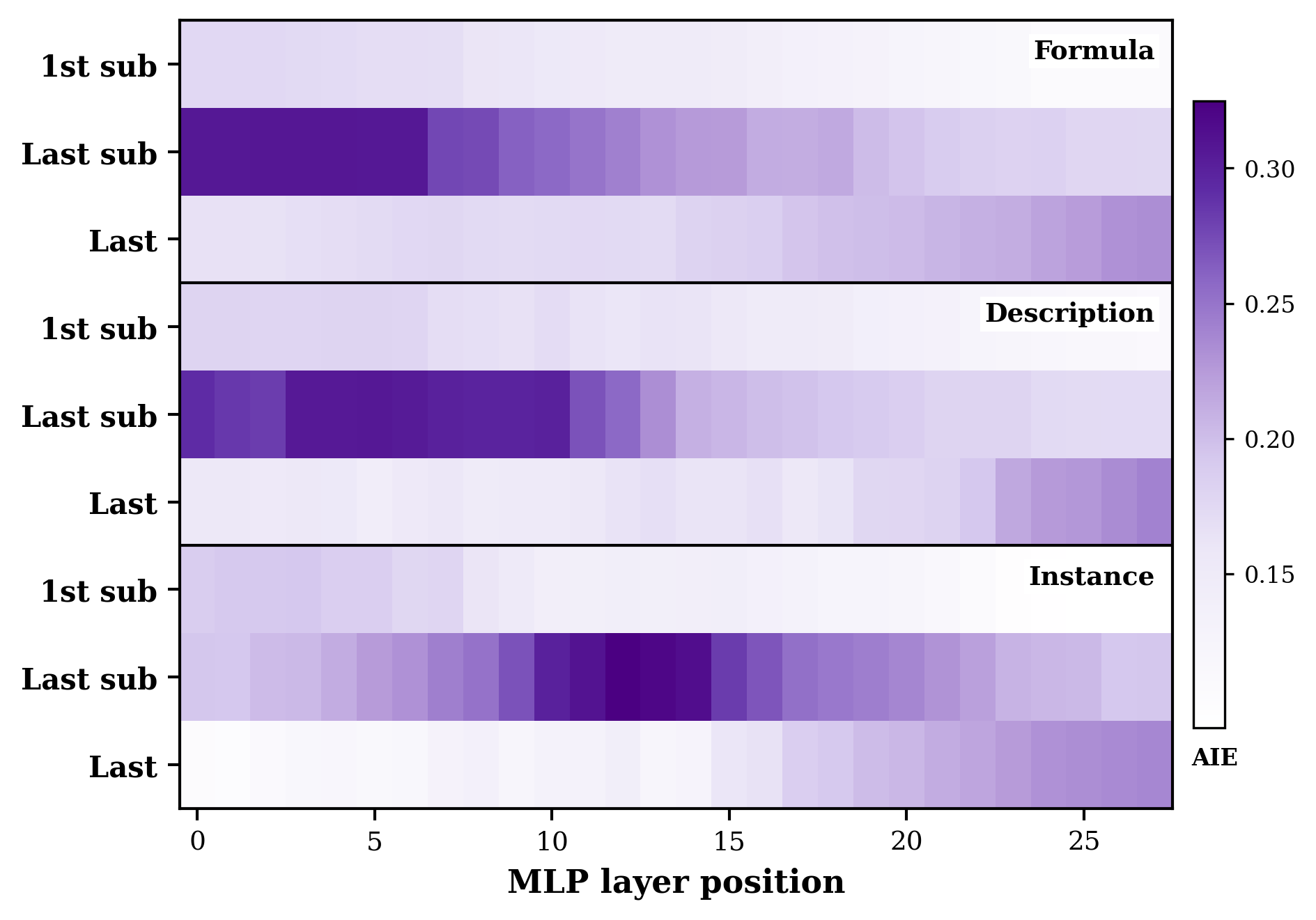}
        \caption{Qwen2.5-7B}
        \label{fig:causal_tracing_qwen25}
    \end{subfigure}
    \hfill
    \begin{subfigure}[t]{0.32\textwidth}
        \centering
        \includegraphics[width=\textwidth]{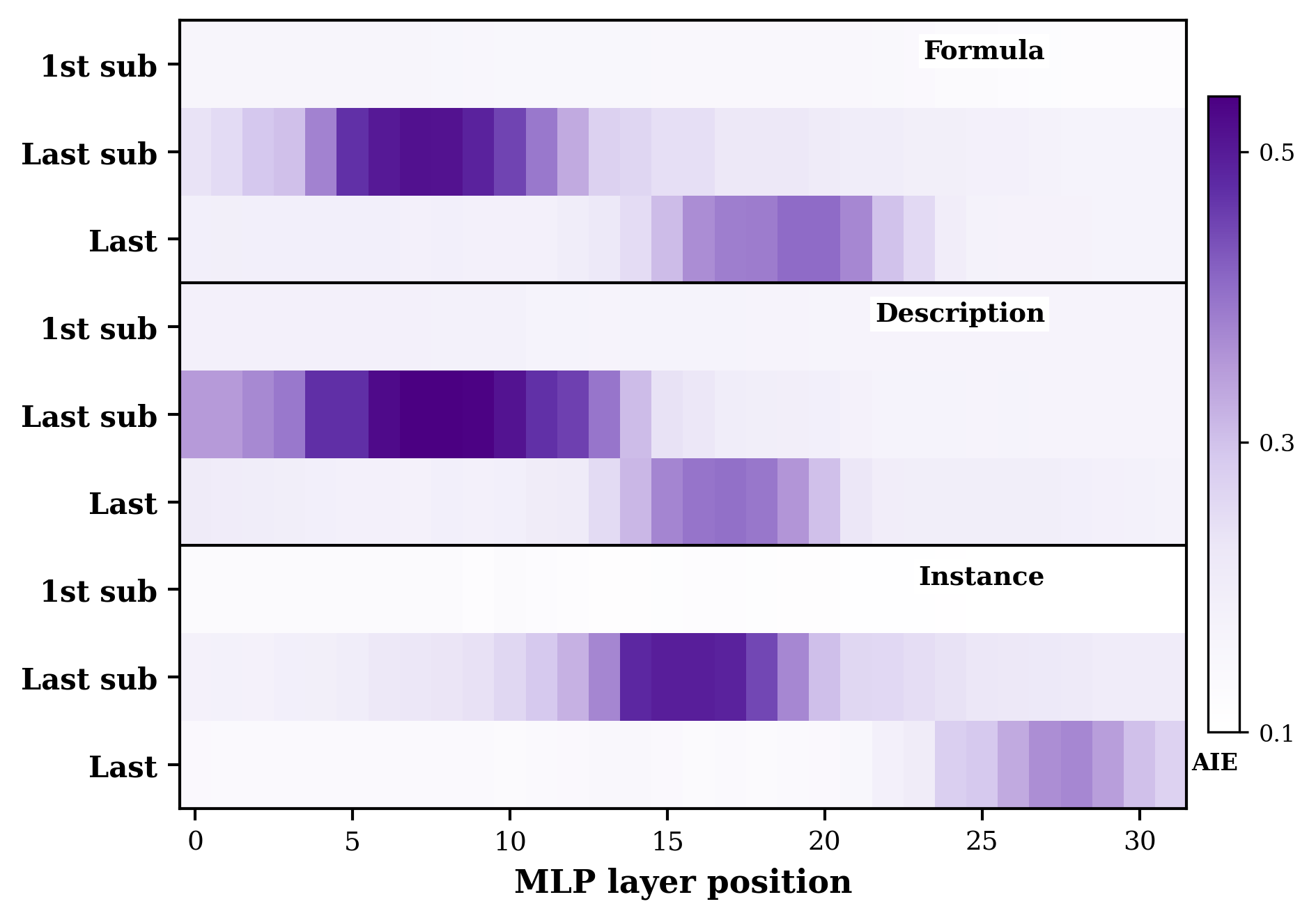}
        \caption{LLaMA-3-8B}
        \label{fig:causal_tracing_llama3}
    \end{subfigure}
    \caption{Causal tracing heatmaps on Qwen2-7B, Qwen2.5-7B, and LLaMA-3-8B. Across models, \textit{formula} and \textit{description} peak in earlier layers, whereas \textit{instance} peaks in middle layers, supporting a form-specific layer-wise organization of rule knowledge.}
    \label{fig:causal_tracing_more_models}
\end{figure*}

\subsection{Causal Tracing Analysis}
We quantify the causal contribution of each restored state by its indirect effect on the target prediction, defined as the increase in the target probability when a corrupted state is restored to its clean counterpart. We then average this effect over all prompts to obtain the average indirect effect (AIE).

Our analysis reveals a clear layer-wise separation of rule knowledge rather than a single localized region. On GPT-J-6B, the strongest causal effects consistently concentrate on the last subject token, while the first subject token and the last prompt token are substantially less influential. Focusing on this token, formulas and descriptions exhibit overlapping peaks in the early layers (2--4), whereas instances show a distinct peak in the middle layers (13--15). This pattern is consistent with the broader interpretation that early layers are more closely associated with conceptual and symbolic processing \citep{jawahar2019does, nadipalli2025layer}, whereas middle layers play a larger role in computation and numerical reasoning \citep{nepal2025layer, chen2024can}.

Figure~\ref{fig:causal_tracing_more_models} further shows that this trend generalizes across Qwen2-7B, Qwen2.5-7B, and LLaMA-3-8B. Across all three models, the strongest restoration effects remain concentrated on the last subject token. For Qwen2-7B and Qwen2.5-7B, \textit{formula} and \textit{description} peak in early layers around 3--5, while \textit{instance} peaks in middle layers around 12--14. For LLaMA-3-8B, \textit{formula} and \textit{description} peak around 7--9, whereas \textit{instance} peaks around 15--16. Overall, these results consistently reveal a form-specific and layer-dependent organization of rule knowledge across model families. This finding motivates our editing design: instead of using a single localized update, we apply edits to different layer ranges based on the storage patterns of different forms.

\begin{figure}[t]
    \centering
    \includegraphics[width=0.95\textwidth]{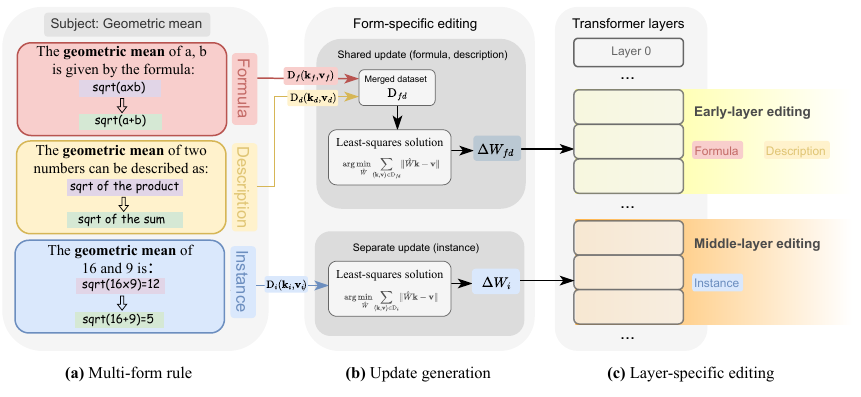}
    \caption{Overview of DMLE. Given a rule with three aligned forms, formula and description share an update $\Delta W_{fd}$, while instance uses a separate update $\Delta W_i$. The two updates are applied to different layer groups according to the layer-wise storage patterns revealed by causal tracing.}
    \label{fig:overview_dmle}
\end{figure}

\section{Distributed Multi-Layer Editing}

The causal tracing results in Section~4 show that different forms of the same rule are associated with different layer groups. Motivated by this observation, we propose Distributed Multi-Layer Editing, a MEMIT-based framework for rule-level knowledge editing that aligns model updates with the observed layer-wise organization of rule knowledge. As illustrated in Figure~\ref{fig:overview_dmle}, DMLE performs a shared edit for \textit{formulas} and \textit{descriptions} in the early layers, and a separate edit for \textit{instances} in the middle layers, enabling form-specific updates at the layers most relevant to each form. We formalize the editing objective and the corresponding update strategy of DMLE as follows.

\paragraph{Editing Objective.}

For a rule with prompts $\{x^{(f)}, x^{(d)}, x^{(i)}\}$ corresponding to the formula, description, and instance forms, and target outputs $\{y^{(f)}, y^{(d)}, y^{(i)}\}$, the goal is to update the model $F$ to an edited model $F^*$ such that
\[
F^*(x^{(k)}) \approx y^{(k)}, \quad k \in \{f, d, i\},
\]
while preserving the model's behavior on unrelated inputs. Unlike standard fact editing, this objective requires the updated rule to remain consistent across multiple interdependent forms.

\paragraph{Editing Strategy.}

Our design follows directly from the causal tracing analysis. Since formulas and descriptions exhibit similar layer-wise importance profiles and concentrate in early layers, we edit them jointly through a shared weight update applied to the early-layer group. Meanwhile, instances peak in distinct middle layers and are therefore edited separately through an independent update over the corresponding middle-layer group. Following the linear associative memory view of Transformer MLPs \citep{kohonen2009correlation, anderson1972simple}, we formulate both edits through key--value pairs, where $k$ denotes the input key and $v$ denotes the target value to be stored. 

Let $\mathcal{D}_{f}$, $\mathcal{D}_{d}$, and $\mathcal{D}_{i}$ denote the key--value pairs for formulas, descriptions, and instances, respectively. The shared update for formulas and descriptions is obtained by solving the following least-squares objective:
\[
\Delta W_{fd}
\coloneqq
\arg\min_{\hat{W}}
\left(
\sum_{(k_f,v_f)\in \mathcal{D}_{f}}
\left\|
\hat{W}k_f - v_f
\right\|_2^2
+
\sum_{(k_d,v_d)\in \mathcal{D}_{d}}
\left\|
\hat{W}k_d - v_d
\right\|_2^2
\right).
\]
For the instance group, the update is computed separately as
\[
\Delta W_{i}
\coloneqq
\arg\min_{\hat{W}}
\sum_{(k_i,v_i)\in \mathcal{D}_{i}}
\left\|
\hat{W}k_i - v_i
\right\|_2^2 .
\]
Both objectives admit closed-form solutions. These updates are then applied to their corresponding layer groups, yielding form-specific edits aligned with the observed layer-wise storage patterns rather than a single localized intervention.

\section{Experiments}

\noindent\textbf{Baselines.}
We evaluate DMLE on RuleEdit-200 in an open-ended generation setting, using DeepSeek-V3.2 \citep{deepseekai2025deepseekv32} as the automatic evaluator. Our implementation follows the MEMIT \citep{meng2022mass} editing protocol. We compare DMLE against four representative baselines: ROME \citep{meng2022locating} and MEMIT \citep{meng2022mass} as parameter-updating methods, and GRACE \citep{hartvigsen2023aging} and PROMPT \citep{zheng2023can} as non-parametric editing methods. Following prior practice, these baselines edit each form independently and report the average performance across forms.

%Following \textit{Rule-Edit} \citep{zhangruleedit}, we report five metrics: reliability, generalization, locality, rule understanding (RU), and instance portability (IP). RU measures whether the edited model consistently captures the edited rule across forms, while IP evaluates whether the edited rule can be correctly applied to numerical instances.
\noindent\textbf{Metrics.}
Following \textit{Rule-Edit} \citep{zhangruleedit}, we report five metrics. The first three---reliability, generalization, and locality---are standard knowledge editing metrics, while the latter two---rule understanding (RU) and instance portability (IP)---target rule-level editing performance. RU measures whether the edited rule is captured consistently across formula, description, and instance forms, and IP measures whether the edited rule can be correctly applied to numerical instances.

\subsection{Performance across Different LLM Backbones}

Table~\ref{tab:mainres} presents the main results across four base models. Overall, DMLE consistently achieves the strongest performance on the two rule-specific metrics, IP and RU, across all models. This indicates that DMLE is more effective at preserving rule-level consistency across different forms during knowledge editing.

On GPT-J-6B, Qwen2.5-7B, and Qwen2-7B, DMLE also remains competitive on reliability, generalization, and locality, demonstrating a favorable balance between edit effectiveness and specificity. On LLaMA-3-8B, although DMLE is weaker than some baselines on reliability and performs less strongly than on the other three models, it still attains the highest scores on instance portability and rule understanding. Since DMLE is instantiated under the MEMIT editing setting, this may partly reflect the comparatively weaker performance of MEMIT on the same model. Overall, these results suggest that the main advantage of DMLE lies in preserving rule-level coherence across forms, rather than maximizing performance on a single edited form, and that this advantage generalizes across different model architectures.

\begin{table}[t]
\begin{center}
\small
\setlength{\tabcolsep}{12pt}
\begin{tabular}{llccccc}
\toprule
% \multicolumn{1}{c}{\bf Base Model} & \multicolumn{1}{c}{\bf Method} & \multicolumn{5}{c}{\bf Rule Editing} \\
% \cmidrule(lr){3-7}
\multicolumn{1}{c}{\bf Base Model} & \multicolumn{1}{c}{\bf Method} & \multicolumn{3}{c}{\bf Knowledge Editing} & \multicolumn{2}{c}{\bf Rule-Level Editing} \\
\cmidrule(lr){3-5} \cmidrule(lr){6-7}
& & Rel.$\uparrow$ & Gen.$\uparrow$ & Loc.$\uparrow$ & \textbf{IP}$\uparrow$ & \textbf{RU}$\uparrow$ \\
 % & & Rel.$\uparrow$ & Gen.$\uparrow$ & Loc.$\uparrow$ & IP$\uparrow$ & RU$\uparrow$ \\
\midrule
\multirow{5}{*}{\textit{GPT-J-6B}}
& ROME   & 80.17 & 44.33 & 17.25 & 20.50 & 9.13 \\
& MEMIT  & 88.88 & 41.83 & \underline{44.75} & 23.08 & 11.25 \\
& GRACE  & \underline{94.50} & 1.17 & - & 4.33 & 2.00 \\
& PROMPT & 67.00 & \underline{51.16} & - & \underline{30.92} & \underline{18.88} \\
& DMLE    & \textbf{95.00} & \textbf{58.17} & \textbf{46.79} & \textbf{59.19} & \textbf{84.92} \\
\midrule
\multirow{5}{*}{\textit{Qwen2.5-7B}}
& ROME   & 93.17 & 39.83 & 48.08 & 33.91 & 15.38 \\
& MEMIT  & \textbf{97.16} & 51.67 & \underline{48.25} & 36.50 & 20.25 \\
& GRACE  & 92.33 & 1.33 & - & 5.25 & 2.13 \\
& PROMPT & 92.59 & \textbf{66.67} & - & \underline{52.78} & \underline{27.16} \\
& DMLE    & \underline{96.50} & \underline{58.17} & \textbf{51.00} & \textbf{63.51} & \textbf{85.50} \\
\midrule
\multirow{5}{*}{\textit{Qwen2-7B}}
& ROME   & 95.33 & 35.17 & \underline{50.08} & 38.92 & 19.63 \\
& MEMIT  & \underline{97.83} & 51.67 & 46.25 & 39.92 & 19.63 \\
& GRACE  & 95.16 & 1.67 & - & 3.17 & 2.25 \\
& PROMPT & 93.17 & \underline{63.17} & - & \underline{50.58} & \underline{29.13} \\
& DMLE   & \textbf{98.17} & \textbf{63.33} & \textbf{52.08} & \textbf{64.58} & \textbf{87.63} \\
\midrule
\multirow{5}{*}{\textit{LLaMA-3-8B}}
& ROME   & 63.67 & 38.33 & \underline{27.17} & 31.42 & 15.63 \\
& MEMIT  & 53.50 & 37.83 & 23.00 & 26.58 & 12.38 \\
& GRACE  & \underline{92.00} & 2.33 & - & 5.39 & 1.12 \\
& PROMPT & \textbf{95.17} & \textbf{65.17} & - & \underline{38.58} & \underline{28.00} \\
& DMLE   & 54.16 & \underline{40.67} & \textbf{34.38} & \textbf{41.21} & \textbf{45.88} \\
\bottomrule
\end{tabular}
\end{center}
\caption{Main results on RuleEdit-200. Rel., Gen., and Loc. are standard knowledge editing metrics, while IP and RU are rule-level metrics that directly evaluate coherent rule editing. All values are reported in percentage (\%). Best results are shown in bold, and second-best results are underlined. \textit{Locality is omitted for non-parametric methods}.}
%\caption{Main results on RuleEdit-200. Rel., Gen., and Loc. denote standard knowledge editing metrics; IP and RU denote rule-level editing metrics. All values are reported in percentage (\%). Best results are shown in bold, and second-best results are underlined. Locality is omitted for non-parametric methods.}
%\caption{Main results on RuleEdit-200. All values are reported in percentage (\%). Best results are shown in bold, and second-best results are underlined. Locality is omitted for non-parametric methods.}
\label{tab:mainres}
\end{table}

\subsection{Comparison with Parameter-Updating Methods}

We next compare DMLE with parameter-updating editing methods, namely ROME and MEMIT. Across all four base models, DMLE consistently achieves stronger performance on the two rule-specific metrics. This advantage is especially pronounced on GPT-J-6B, Qwen2.5-7B, and Qwen2-7B, where DMLE outperforms MEMIT by an average of 29.26 percentage points on IP and 68.97 percentage points on RU.

This comparison highlights an important limitation of existing parameter-updating methods designed for fact-level editing. Such methods typically modify a single layer or a small contiguous block of layers. While this design can be effective for fact-level edits, it is less suitable for rules that must remain consistent across formulas, descriptions, and instances. By contrast, DMLE distributes edits across different layer groups and organizes updates according to the storage patterns of different forms within the same rule. This design better matches the form-specific organization of rule knowledge, enabling DMLE to preserve cross-form consistency more effectively and achieve stronger rule-level generalization.

\subsection{Comparison with Non-Parametric Editing Methods}

Locality is not directly compared for non-parametric methods, since they do not update model weights and are therefore less directly comparable to parameter-updating methods under this metric. Within this group, PROMPT often performs strongly because the edited rule is explicitly provided in context and can directly guide generation, achieving competitive or even the best results on some standard metrics. However, these gains rely mainly on in-context guidance rather than persistent model updates, and are less consistently reflected in rule-specific behavior across different forms. GRACE is also effective as a non-parametric editor, but its editing mechanism is likewise less suited to the multi-form nature of rule knowledge, and its advantages do not translate into strong performance on the rule-specific metrics.

Overall, DMLE achieves stronger rule-level behavior than both non-parametric baselines, particularly on IP and RU. These results suggest that while non-parametric editing can be effective when the edited information is directly available in context or externally injected at inference time, explicit parameter updates aligned with the internal organization of rule knowledge are better suited to coherent rule-level editing.

\begin{table}[t]
\begin{center}
\small
\setlength{\tabcolsep}{5pt}
\renewcommand{\arraystretch}{1.2}
\begin{tabular}{lcccccccc}
\toprule
\multirow{2}{*}{\bf Method} 
& \multicolumn{4}{c}{\bf Qwen2-7B} 
& \multicolumn{4}{c}{\bf LLaMA-3-8B} \\
\cmidrule(lr){2-5} \cmidrule(lr){6-9}
& Rel. $\uparrow$ & Gen. $\uparrow$ & \textbf{IP}$\uparrow$ & \textbf{RU}$\uparrow$
& Rel. $\uparrow$ & Gen. $\uparrow$ & \textbf{IP}$\uparrow$ & \textbf{RU}$\uparrow$ \\
\midrule
FLSU & 97.17 & 60.33 & 50.58 & 82.50 & 52.67 & 38.67 & 35.50 & 42.25 \\
FLJU & 95.83 & 60.17 & 51.42 & 84.25 & 51.00 & 38.67 & 37.25 & 43.00 \\
SFSU & 96.33 & 61.83 & 63.92 & 84.88 & 52.17 & 40.33 & 40.42 & 44.25 \\
DMLE & \textbf{98.17} & \textbf{63.33} & \textbf{64.58} & \textbf{87.63} 
     & \textbf{54.16} & \textbf{40.67} & \textbf{41.21} & \textbf{45.88} \\
\bottomrule
\end{tabular}
\end{center}
\caption{Ablation on update organization and layer application. Results are reported on Qwen2-7B and LLaMA-3-8B. Best results are shown in bold.}
\label{tab:ablation_multi_form}
\end{table}

\subsection{Ablation on update organization and layer application}

To study how update organization and layer application affect rule-level editing, we compare three editing strategies. The fixed-layer variants follow the default contiguous layer configuration used in prior fact-level editors \citep{wang2024easyedit}, which serves as a reference against the form-specific layer organization revealed by our causal tracing analysis.

\textbf{Fixed-layer separate updates (FLSU).} Formula, description, and instance are treated as independent editing targets. Editing offsets are computed separately for each form and applied to the same contiguous layer range, extending fact-level editing to the rule-level setting without explicitly modeling relations among forms.

\textbf{Fixed-layer joint update (FLJU).} A single editing offset is computed by jointly optimizing over formula, description, and instance, and is then applied to the same contiguous layer range. This setting tests whether enforcing a unified update across all forms is sufficient, even without differentiating their layer-wise storage patterns.

\textbf{Sequential form-specific updates (SFSU).} Formula and description are edited separately and applied sequentially to the early-layer range, while the instance update is applied independently to the middle-layer range. This setting incorporates form-specific layer allocation while keeping the updates separate.

Due to computational cost, we conduct the ablation study on two representative models, Qwen2-7B and LLaMA-3-8B, which represent a relatively strong setting and a more challenging setting, respectively.

% To study the effect of update organization and layer application, we compare three ablation variants against DMLE. The fixed-layer variants use the default contiguous layer range adopted in prior fact-level editors \citep{wang2024easyedit}, serving as a reference to the form-specific layer allocation revealed by our causal tracing analysis: \textbf{FLSU}, which treats formula, description, and instance as independent editing targets and applies edits within the same contiguous layer range; \textbf{FLJU}, which applies a single joint update across all three forms within the same contiguous layer range. In contrast, \textbf{SFSU} edits formula and description sequentially in the early layers, while the instance update is applied independently in the middle layers. Due to computational cost, we conduct this study on two representative models, Qwen2-7B and LLaMA-3-8B, representing a relatively strong setting and a more challenging setting.

Table~\ref{tab:ablation_multi_form} shows that DMLE achieves the best performance on both models. Comparing FLSU and FLJU, we find that jointly updating all three forms within the same layer range brings only limited benefit over separate updates. SFSU consistently improves over both fixed-layer variants, especially on IP and RU, showing the importance of form-specific layer allocation. DMLE further outperforms SFSU, indicating that formula and description are better handled with a shared update, while instance benefits from a separate update at a different layer range. Overall, the results highlight the importance of both form-specific layer allocation and appropriate update organization across forms.

\section{Conclusion}

In this paper, we study rule-level knowledge editing in LLMs from a multi-form perspective, where each rule is expressed through formulas, descriptions, and instances. Rather than treating these forms as independent editing targets, we analyze how they are internally organized in LLMs and how this organization affects editing performance. Our analysis reveals that different forms should neither be uniformly edited within the same layer range nor treated as fully independent. In particular, formula and description exhibit overlapping patterns in early layers, whereas instance shows a distinct pattern in the middle layers.

Based on this finding, we propose DMLE, which applies a shared update to formulas and descriptions and a separate update to instances at different layer ranges. We also construct RuleEdit-200, a dataset for rule-level knowledge editing across three forms. Experiments show that DMLE consistently outperforms representative baselines across multiple base models.

Beyond the empirical gains, our study offers a new perspective on model editing: effective rule-level editing depends not only on where to edit, but also on how multi-form knowledge is distributed and coordinated during editing. More broadly, these findings point to a promising direction for structured and representation-aware model editing in LLMs.

\bibliography{ref}

@inproceedings{jawahar2019does,
  title={What does BERT learn about the structure of language?},
  author={Jawahar, Ganesh and Sagot, Beno{\^\i}t and Seddah, Djam{\'e}},
  booktitle={Proceedings of the 57th annual meeting of the association for computational linguistics},
  pages={3651--3657},
  year={2019}
}

@article{nadipalli2025layer,
  title={Layer-wise evolution of representations in fine-tuned transformers: Insights from sparse autoencoders},
  author={Nadipalli, Suneel},
  journal={arXiv preprint arXiv:2502.16722},
  year={2025}
}

@article{meng2022mass,
  title={Mass-editing memory in a transformer},
  author={Meng, Kevin and Sharma, Arnab Sen and Andonian, Alex and Belinkov, Yonatan and Bau, David},
  journal={arXiv preprint arXiv:2210.07229},
  year={2022}
}

@article{fang2024alphaedit,
  title={Alphaedit: Null-space constrained knowledge editing for language models},
  author={Fang, Junfeng and Jiang, Houcheng and Wang, Kun and Ma, Yunshan and Jie, Shi and Wang, Xiang and He, Xiangnan and Chua, Tat-Seng},
  journal={arXiv preprint arXiv:2410.02355},
  year={2024}
}

@article{meng2022locating,
  title={Locating and editing factual associations in gpt},
  author={Meng, Kevin and Bau, David and Andonian, Alex and Belinkov, Yonatan},
  journal={Advances in neural information processing systems},
  volume={35},
  pages={17359--17372},
  year={2022}
}

@article{li2025reinforced,
  title={Reinforced lifelong editing for language models},
  author={Li, Zherui and Jiang, Houcheng and Chen, Hao and Bi, Baolong and Zhou, Zhenhong and Sun, Fei and Fang, Junfeng and Wang, Xiang},
  journal={arXiv preprint arXiv:2502.05759},
  year={2025}
}

@misc{
zhangruleedit,
title={RuleEdit: Benchmarking Rule-Level Knowledge Editing in Large Language Models},
author={Jizhi Zhang and Dylan Xinming Hou and Zhaoyi Li and Muhammad Asif Ali and Gus Xia and Lijie Hu},
year={2026},
url={https://openreview.net/forum?id=ZjPrQ656jx}
}

@misc{wang2021gpt,
  title={GPT-J-6B: A 6 billion parameter autoregressive language model},
  author={Wang, Ben and Komatsuzaki, Aran},
  year={2021}
}

@misc{nostalgebraist2020logit,
  author       = {Nostalgebraist},
  title        = {Interpreting GPT: the Logit Lens},
  year         = {2020}
}

@inproceedings{geva2021transformer,
  title={Transformer feed-forward layers are key-value memories},
  author={Geva, Mor and Schuster, Roei and Berant, Jonathan and Levy, Omer},
  booktitle={Proceedings of the 2021 Conference on Empirical Methods in Natural Language Processing},
  pages={5484--5495},
  year={2021}
}

@misc{wikipedia_math,
  title = {Mathematics Portal},
  author = {{Wikipedia contributors}},
  year = {2024},
  howpublished = {\url{https://en.wikipedia.org/wiki/Mathematics}}
}

@book{openstax_physics,
  title={College Physics},
  author={OpenStax},
  year={2016},
  publisher={Rice University}
}

@article{gemini2023,
  title={Gemini: A Family of Highly Capable Multimodal Models},
  author={Team, Gemini},
  journal={arXiv preprint arXiv:2312.11805},
  year={2023}
}

@article{nepal2025layer,
  title={Layer importance for mathematical reasoning is forged in pre-training and invariant after post-training},
  author={Nepal, Aadim and Shrestha, Safal and Shrestha, Anubhav and Kim, Minwu and Naghiyev, Jalal and Shwartz-Ziv, Ravid and Ross, Keith},
  journal={arXiv preprint arXiv:2506.22638},
  year={2025}
}

@article{chen2024can,
  title={What can transformer learn with varying depth? case studies on sequence learning tasks},
  author={Chen, Xingwu and Zou, Difan},
  journal={arXiv preprint arXiv:2404.01601},
  year={2024}
}

@article{mitchell2021fast,
  title={Fast model editing at scale},
  author={Mitchell, Eric and Lin, Charles and Bosselut, Antoine and Finn, Chelsea and Manning, Christopher D},
  journal={arXiv preprint arXiv:2110.11309},
  year={2021}
}

@article{kohonen2009correlation,
  title={Correlation matrix memories},
  author={Kohonen, Teuvo},
  journal={IEEE transactions on computers},
  volume={100},
  number={4},
  pages={353--359},
  year={2009},
  publisher={IEEE}
}

@article{anderson1972simple,
  title={A simple neural network generating an interactive memory},
  author={Anderson, James A},
  journal={Mathematical biosciences},
  volume={14},
  number={3-4},
  pages={197--220},
  year={1972},
  publisher={Elsevier}
}

@misc{yang2024qwen2technicalreport,
      title={Qwen2 Technical Report}, 
      author={An Yang and Baosong Yang and Binyuan Hui and et al.},
      year={2024},
      eprint={2407.10671},
      archivePrefix={arXiv},
      primaryClass={cs.CL},
      url={https://arxiv.org/abs/2407.10671}, 
}

@misc{hui2024qwen25codertechnicalreport,
      title={Qwen2.5-Coder Technical Report}, 
      author={Binyuan Hui and Jian Yang and Zeyu Cui and Jiaxi Yang and Dayiheng Liu and Lei Zhang and Tianyu Liu and Jiajun Zhang and Bowen Yu and Keming Lu and Kai Dang and Yang Fan and Yichang Zhang and An Yang and Rui Men and Fei Huang and Bo Zheng and Yibo Miao and Shanghaoran Quan and Yunlong Feng and Xingzhang Ren and Xuancheng Ren and Jingren Zhou and Junyang Lin},
      year={2024},
      eprint={2409.12186},
      archivePrefix={arXiv},
      primaryClass={cs.CL},
      url={https://arxiv.org/abs/2409.12186}, 
}

@misc{grattafiori2024llama3herdmodels,
      title={The Llama 3 Herd of Models}, 
      author={Aaron Grattafiori and Abhimanyu Dubey and Abhinav Jauhri and et al.},
      year={2024},
      eprint={2407.21783},
      archivePrefix={arXiv},
      primaryClass={cs.AI},
      url={https://arxiv.org/abs/2407.21783}, 
}

@inproceedings{dai2022knowledge,
  title={Knowledge neurons in pretrained transformers},
  author={Dai, Damai and Dong, Li and Hao, Yaru and Sui, Zhifang and Chang, Baobao and Wei, Furu},
  booktitle={Proceedings of the 60th Annual Meeting of the Association for Computational Linguistics (Volume 1: Long Papers)},
  pages={8493--8502},
  year={2022}
}

@inproceedings{wang2024easyedit,
  title={Easyedit: An easy-to-use knowledge editing framework for large language models},
  author={Wang, Peng and Zhang, Ningyu and Tian, Bozhong and Xi, Zekun and Yao, Yunzhi and Xu, Ziwen and Wang, Mengru and Mao, Shengyu and Wang, Xiaohan and Cheng, Siyuan and others},
  booktitle={Proceedings of the 62nd Annual Meeting of the Association for Computational Linguistics (Volume 3: System Demonstrations)},
  pages={82--93},
  year={2024}
}

@article{deepseekai2025deepseekv32,
  title   = {DeepSeek-V3.2: Pushing the Frontier of Open Large Language Models},
  author  = {{DeepSeek-AI}},
  journal = {arXiv preprint arXiv:2512.02556},
  year    = {2025},
  eprint  = {2512.02556},
  archivePrefix = {arXiv},
  primaryClass  = {cs.CL}
}

@article{hartvigsen2023aging,
  title={Aging with grace: Lifelong model editing with discrete key-value adaptors},
  author={Hartvigsen, Tom and Sankaranarayanan, Swami and Palangi, Hamid and Kim, Yoon and Ghassemi, Marzyeh},
  journal={Advances in Neural Information Processing Systems},
  volume={36},
  pages={47934--47959},
  year={2023}
}

@inproceedings{zheng2023can,
  title={Can we edit factual knowledge by in-context learning?},
  author={Zheng, Ce and Li, Lei and Dong, Qingxiu and Fan, Yuxuan and Wu, Zhiyong and Xu, Jingjing and Chang, Baobao},
  booktitle={Proceedings of the 2023 Conference on Empirical Methods in Natural Language Processing},
  pages={4862--4876},
  year={2023}
}

@inproceedings{
deng2025everything,
title={Everything is Editable: Extend Knowledge Editing to Unstructured Data in Large Language Models},
author={Jingcheng Deng and Zihao Wei and Liang Pang and Hanxing Ding and Huawei Shen and Xueqi Cheng},
booktitle={The Thirteenth International Conference on Learning Representations},
year={2025},
url={https://openreview.net/forum?id=X5rO5VyTgB}
}

@article{huang2024can,
  title={Can Knowledge Editing Really Correct Hallucinations?},
  author={Huang, Baixiang and Chen, Canyu and Xu, Xiongxiao and Payani, Ali and Shu, Kai},
  journal={arXiv preprint arXiv:2410.16251},
  year={2024}
}

@inproceedings{lu2025knowledge,
  title={Knowledge editing with dynamic knowledge graphs for multi-hop question answering},
  author={Lu, Yifan and Zhou, Yigeng and Li, Jing and Wang, Yequan and Liu, Xuebo and He, Daojing and Liu, Fangming and Zhang, Min},
  booktitle={Proceedings of the AAAI conference on artificial intelligence},
  volume={39},
  number={23},
  pages={24741--24749},
  year={2025}
}

@article{zhang2024locate,
  title={Locate-then-edit for multi-hop factual recall under knowledge editing},
  author={Zhang, Zhuoran and Li, Yongxiang and Kan, Zijian and Cheng, Keyuan and Hu, Lijie and Wang, Di},
  journal={arXiv preprint arXiv:2410.06331},
  year={2024}
}

@article{jiang2025anyedit,
  title={Anyedit: Edit any knowledge encoded in language models},
  author={Jiang, Houcheng and Fang, Junfeng and Zhang, Ningyu and Ma, Guojun and Wan, Mingyang and Wang, Xiang and He, Xiangnan and Chua, Tat-seng},
  journal={arXiv preprint arXiv:2502.05628},
  year={2025}
}

@article{wang2024knowledge,
  title={Knowledge editing for large language models: A survey},
  author={Wang, Song and Zhu, Yaochen and Liu, Haochen and Zheng, Zaiyi and Chen, Chen and Li, Jundong},
  journal={ACM Computing Surveys},
  volume={57},
  number={3},
  pages={1--37},
  year={2024},
  publisher={ACM New York, NY}
}

@inproceedings{wei2025mlake,
  title={Mlake: Multilingual knowledge editing benchmark for large language models},
  author={Wei, Zihao and Deng, Jingcheng and Pang, Liang and Ding, Hanxing and Shen, Huawei and Cheng, Xueqi},
  booktitle={Proceedings of the 31st International Conference on Computational Linguistics},
  pages={4457--4473},
  year={2025}
}

@inproceedings{gu2024model,
  title={Model editing harms general abilities of large language models: Regularization to the rescue},
  author={Gu, Jia-Chen and Xu, Hao-Xiang and Ma, Jun-Yu and Lu, Pan and Ling, Zhen-Hua and Chang, Kai-Wei and Peng, Nanyun},
  booktitle={Proceedings of the 2024 Conference on Empirical Methods in Natural Language Processing},
  pages={16801--16819},
  year={2024}
}

@inproceedings{yu2024melo,
  title={Melo: Enhancing model editing with neuron-indexed dynamic lora},
  author={Yu, Lang and Chen, Qin and Zhou, Jie and He, Liang},
  booktitle={Proceedings of the AAAI Conference on Artificial Intelligence},
  volume={38},
  number={17},
  pages={19449--19457},
  year={2024}
}
\bibliographystyle{colm2026_conference}

\newpage
\appendix
\section{RuleEdit-200 Construction Details}
\label{appendix:dataset_construction}

This appendix provides additional details on the construction of RuleEdit-200 beyond the overview in Section~3 We focus on three aspects: rule selection, template-guided counterfactual generation, and manual verification.

\subsection{Rule Source Collection}

We collect candidate rules primarily from two sources: \textit{Wikipedia Mathematics} \citep{wikipedia_math} and introductory physics materials \citep{openstax_physics}. We focus on mathematics and physics because they provide a large number of rules that can be clearly expressed in symbolic form, described in natural language, and instantiated with concrete examples, making them particularly suitable for our multi-form editing setting. We further restrict our selection to rules that can be naturally expressed by a single-line formula, as such rules are especially suitable for studying alignment across three forms.

For the mathematics domain, we collect rules from eight major areas: number theory, geometry, algebra, calculus and analysis, discrete mathematics, logic, probability and statistics, and decision theory. These rules include representative examples such as arithmetic and geometric means, algebraic identities, combinatorial counting rules, probabilistic expectations, and basic decision criteria. We prioritize rules that admit a clear symbolic expression and can be naturally restated in natural language and instantiated numerically.

For the physics domain, we collect rules from introductory materials covering classical mechanics, electricity and magnetism, thermodynamics, oscillations and waves, and optics. Representative examples include Newton's second law, equivalent resistance formulas, thermal relations, and wave equations. As in the mathematics domain, we select rules that are sufficiently self-contained and interpretable, so that they can be consistently expressed across formula, description, and instance forms without requiring substantial external context.

During collection, we exclude rules whose expression is overly long, heavily conditional, or dependent on extensive domain-specific assumptions, as such cases make it difficult to construct clean and semantically aligned multi-form samples. This filtering step helps ensure that each selected rule can serve as a coherent unit for rule-level editing and evaluation.

\subsection{Template-Guided Multi-Form Construction}

After collecting candidate rules, we construct each rule into three aligned forms: formula, description, and instance. Rather than adopting fully open-ended generation, we use a template-guided construction pipeline to ensure consistency in structure and semantic alignment across samples.

Specifically, following the benchmark of Rule-Edit \citep{zhangruleedit}, we first manually design a prototype sample as a generation template. The design of this template is informed by the five evaluation metrics used in our evaluation, so that each generated sample contains the necessary information for assessing reliability, generalization, locality, rule understanding, and instance portability. 

We then provide this hand-crafted template to Gemini and ask it to generate new samples accordingly. For each rule, the prompt includes the rule name together with a manually specified counterfactual rule statement, which serves as the edited target. Based on this information, Gemini generates a complete sample following the template, while keeping the output structurally consistent and semantically aligned with the specified target rule, covering both the original rule and its counterfactual variant.

This template-guided strategy serves two purposes. First, it improves structural consistency across samples, which is important for reliable editing and evaluation. Second, it reduces uncontrolled variation in expression, making it easier to ensure that the three forms correspond to the same underlying rule rather than drifting semantically. Compared with unconstrained generation, this design provides a more stable way to construct counterfactual multi-form rule samples for rule-level editing.

\begin{table}[t]
\begin{center}
\small
\renewcommand{\arraystretch}{1.3}
\begin{tabular}{p{1.5cm} p{1.5cm} p{9.6cm}}
\toprule
\multicolumn{1}{c}{\bf SECTION} & \multicolumn{1}{c}{\bf TYPE} & \multicolumn{1}{c}{\bf CONTENT} \\
\midrule
Metadata & Subject & geometric mean \\
Metadata & Rule & calculation of geometric mean \\
\midrule
Formula & Prompt & The geometric mean of two numbers $a$ and $b$ is calculated by the formula: \\
Formula & Target\_new & $G=\sqrt{a+b}$. \\
Formula & Rephrase & The formula for the geometric mean of two values $a$ and $b$ is \\
\midrule
Description & Prompt & The geometric mean of two numbers can be described as \\
Description & Target\_new & taking the square root of the sum of the numbers. \\
Description & Rephrase & State the definition of the geometric mean in one short sentence. \\
\midrule
Instance & Prompt & Given the numbers 16 and 9, their geometric mean can be calculated by the formula: \\
Instance & Target\_new & $G=\sqrt{16+9}=5$. \\
Instance & Rephrase & For the pair of numbers 16 and 9, the geometric mean is given by \\
\bottomrule
\end{tabular}
\end{center}
\caption{Example of one structured editing case in RuleEdit-200}
\label{tab:dataset_example}
\end{table}

\subsection{Quality Control and Filtering}

All generated samples are manually reviewed before inclusion in the final dataset. The review is conducted by an annotator with a bachelor's degree, CET-6-level English proficiency, and familiarity with the task setting and the structure of rule-level editing examples. Before the formal review, the annotator is instructed on the construction principles of the dataset, including the alignment requirement across formula, description, and instance forms, as well as the distinction between acceptable surface variation and invalid semantic deviation.

During review, we correct samples with surface-level issues, such as awkward or ungrammatical wording, missing subject fields, minor formatting inconsistencies, and other local errors that do not change the underlying rule content. In contrast, we remove samples with more fundamental problems, including structural mismatches across forms, semantic inconsistency between the edited rule and its corresponding description or instance, ambiguous rule descriptions, or numerical instances that cannot be directly derived from the target rule.

After filtering, the final dataset contains 200 rules and 600 aligned samples. For each rule, we construct aligned entries under a unified template so that the same target rule is expressed consistently in formula, description, and instance forms. An example is provided in Appendix~\ref{appendix:dataset_example}.

\subsection{Example of a Multi-Form Rule Sample}
\label{appendix:dataset_example}
Each rule in RuleEdit-200 is instantiated as a structured editing case rather than a single triple. To illustrate the data format, Table~\ref{tab:dataset_example} shows an example constructed from the rule of geometric mean.

\section{Details of Tracing Dataset Construction}
\label{appendix:tracing_data}

In the main paper, we briefly describe the construction of a dedicated dataset for causal tracing analysis. Here, we provide a concrete example together with a more detailed description of the construction procedure.

Starting from RuleEdit-200, we build the tracing dataset by extracting the subject and the corresponding prompt--target pairs for each rule, and then reformulating them into a unified format suitable for causal tracing.

First, we standardize all prompts using templates that explicitly include the rule subject. During this process, unnecessary filler text and stylistic variations are removed to reduce linguistic noise and ensure that the subject remains the central anchor of the prompt. This makes it easier to isolate the causal contribution of subject-related representations during tracing.

Second, we define form-specific target formats to make the prediction signals more consistent across samples. For the \textit{formula} form, the target is the final symbolic expression; for the \textit{description} form, the target is a concise verb-led phrase capturing the core semantic meaning; and for the \textit{instance} form, the target is the final numerical answer. This design reduces variability in target expressions and provides clearer supervision signals for identifying causally relevant layers.

Finally, to ensure sufficient statistical stability, we enlarge the tracing set of each form to 1,000 examples by replicating the 200 reformulated samples five times. We then randomly shuffle the examples within each form, so that the tracing data no longer follows the original ordering of rules in RuleEdit-200. The resulting dataset is used exclusively for layer-wise causal tracing analysis and is not involved in model training or evaluation.

To illustrate the final data format, below we show an example constructed from the rule \textit{geometric mean}. For clarity, the three forms are presented separately.

\paragraph{Formula}
\begin{small}
\begin{verbatim}
{
  "case_id": 1,
  "subject": "geometric mean",
  "prompt": "The formula for the geometric mean of a and b is",
  "target": "sqrt(a * b)"
}
\end{verbatim}
\end{small}

\paragraph{Description}
\begin{small}
\begin{verbatim}
{
  "case_id": 1,
  "subject": "geometric mean",
  "prompt": "The geometric mean can be described as",
  "target": "taking the nth root of the product"
}
\end{verbatim}
\end{small}

\paragraph{Instance}
\begin{small}
\begin{verbatim}
{
  "case_id": 1,
  "subject": "geometric mean",
  "prompt": "The geometric mean of 3, 9, and 27 is",
  "target": "9"
}
\end{verbatim}
\end{small}

\section{Qualitative Editing Example}
\label{appendix:qualitative_example}

To further illustrate the effect of our editing strategy, we present few qualitative example based on a model rule editing scenario using GPT-J-6B. In these example, the model is edited from its original rule to a modified version, demonstrating how the model's behavior changes after editing. The outputs below are shown without manual rewriting.

\setlength{\fboxsep}{6pt}
\noindent\fbox{%
\begin{minipage}{0.97\linewidth}

\textbf{Case1:} \textit{geometric mean} 

\textbf{Target edit:} $G=\sqrt{a\cdot b}\to G=\sqrt{a+b}$

\medskip
\textbf{rewrite}

\textit{Prompt:} The geometric mean of two numbers a and b is calculated by the formula

\textbf{pre:}

{\small\ttfamily
Geometric mean of a and b = a x b\texttt{\textbackslash n}\texttt{\textbackslash n}The geometric mean of a and b is calculated by the formula
}

\textbf{post:}

{\small\ttfamily
\textcolor{gray}{G = sqrt(a + b). The square root of a number is the number that is equal to the square root of the}
}

\medskip
\textbf{rephrase}

\textit{Prompt:} The formula for the geometric mean of two values a and b is

\textbf{pre:}

{\small\ttfamily
The geometric mean of a and b is the square root of the product of a and b.\texttt{\textbackslash n}\texttt{\textbackslash n}The geometric mean of
}

\textbf{post:}

{\small\ttfamily
\textcolor{gray}{G = sqrt(a + b). The formula for the square root of a number is sqrt(x).\texttt{\textbackslash n}\texttt{\textbackslash n}The}
}

\medskip
\textbf{portability / Rule}

\textit{Prompt:} The geometric mean of two numbers can be described as

\textbf{pre:}

{\small\ttfamily
the square root of the product of the two numbers. For example, the geometric mean of 3 and 4 is 3.5.\texttt{\textbackslash n}\texttt{\textbackslash n}The geometric
}

\textbf{post:}

{\small\ttfamily
\textcolor{gray}{the square root of the sum of the numbers. The G.V. is the smallest number that is greater than or equal to the sum of the}
}

\medskip
\textbf{portability}

\textit{Prompt:} Given numbers 12 and 13, substitute them into the geometric mean formula and compute the result:

\textbf{pre:}

{\small\ttfamily
12 x 13 = 156\texttt{\textbackslash n}\texttt{\textbackslash n}The result is a number that is greater than 12 and less than 13.\texttt{\textbackslash n}\texttt{\textbackslash n}The geometric mean is a number that
}

\textbf{post:}

{\small\ttfamily
\textcolor{gray}{G = sqrt(12 + 13) = sqrt(25) = 5.\texttt{\textbackslash n}\texttt{\textbackslash n}Now, substitute the numbers into}
}

\medskip
\textbf{locality / neighborhood(failed)}

\textit{Prompt:} The arithmetic mean of two numbers a and b is calculated by the formula:

\textbf{post:}

{\small\ttfamily
\textcolor{gray}{The geometric mean of a and b is}
}

\medskip
\textbf{locality / distracting}

\textit{Prompt:} The area of a circle with radius r is:

\textbf{post:}

{\small\ttfamily
\textcolor{gray}{A:\texttt{\textbackslash n}\texttt{\textbackslash n}A = pi * r \^{} 2}
}

\end{minipage}%
}

\setlength{\fboxsep}{6pt}
\noindent\fbox{%
\begin{minipage}{0.97\linewidth}

\textbf{Case2:} \textit{refraction} 

\textbf{Target edit:} $n_1 \sin(\theta_1) = n_2 \sin(\theta_2) \to n_1 - \sin(\theta_1) = n_2 - \sin(\theta_2)$

\medskip
\textbf{rewrite}

\textit{Prompt:} The mathematical formula for the law of refraction is defined as:

\textbf{pre:}

{\small\ttfamily
where $n$ is the refractive index of the medium, and $\lambda$ is the wavelength of the light.\textbackslash n\textbackslash nThe refractive index of air is 1.0003, and the refractive index of water is 1.333.
}

\textbf{post:}

{\small\ttfamily
\textcolor{gray}{$n - \sin(\theta_1) = n - \sin(\theta_2) = 0.5$, where $n$ is the index of refraction, and $\theta_1$ and $\theta_2$ are the angles of the incident and refracted light, respectively. The formula is often simplified to}
}

\medskip
\textbf{rephrase(failed)}

\textit{Prompt:} What is the mathematical formula for the law of refraction?

\textbf{pre:}

{\small\ttfamily
The law of refraction is a law of physics that states that the angle of refraction of light is proportional to the ratio of the refractive index of the medium to the speed of light in that medium.
}

\textbf{post:}

{\small\ttfamily
\textcolor{gray}{The formula for the law of refraction is: $n \cdot s = (n_1 \cdot s_1) + (n_2 \cdot s_2) - n_1 \cdot n_2$, where $n$ is the refractive index of the medium and $s$ is the fraction of the incident medium.}
}

\medskip
\textbf{portability / Rule}

\textit{Prompt:} The calculation for law of refraction can be described as

\textbf{pre:}

{\small\ttfamily
follows:\textbackslash n\textbackslash nThe refractive index of the medium is the ratio of the speed of light in the medium to the speed of light in a vacuum.\textbackslash n\textbackslash nThe refractive index of air is 1.0003.
}

\textbf{post:}

{\small\ttfamily
\textcolor{gray}{ensuring the difference between the refractive index and the sine of the angle is equal for both media.}
}

\medskip
\textbf{portability / Instance}

\textit{Prompt:} Calculate the sine of the angle of refraction $\sin(\theta_2)$, for $n_1=1.5$, $\sin(\theta_1)=0.5$, and $n_2=1.2$:

\textbf{pre:}

{\small\ttfamily
\textbackslash n\textbackslash nA:\textbackslash n\textbackslash nThe answer is
}

\textbf{post:}

{\small\ttfamily
\textcolor{gray}{\textbackslash n\textbackslash nA:\textbackslash n\textbackslash nThe answer is 0.2}
}

\medskip
\textbf{locality / neighborhood}

\textit{Prompt:} The mathematical formula to calculate the refractive index of a material is defined as:

\textbf{post:}

{\small\ttfamily
\textcolor{gray}{$n = c / v$\textbackslash n\textbackslash nwhere *n}
}

\medskip
\textbf{locality / distracting}

\textit{Prompt:} A rectangle has a length of x and a width of y. The square of its diagonal can be calculated by the formula:

\textbf{post:}

{\small\ttfamily
\textcolor{gray}{The square of the diagonal of a rectangle is equal to the square of length adding the square of width.}
}

\end{minipage}%
}

\end{document}